\documentclass{article}

\usepackage{arxiv}

\usepackage[utf8]{inputenc} 
\usepackage[T1]{fontenc}    
\usepackage{hyperref}       
\usepackage{url}            
\usepackage{booktabs}       
\usepackage{amsfonts}       
\usepackage{nicefrac}       
\usepackage{microtype}      
\usepackage{lipsum}
\usepackage{graphicx}
\usepackage{multirow}
\graphicspath{ {./images/} }

\title{Explainable CNN-attention Networks (C-Attention Network) for Automated Detection of Alzheimer's Disease\thanks{This paper has been accepted for publication in BIOKDD 2020.}}

\author{
 Ning Wang \\
  Department of Electrical and Computer Engineering\\
  Stevens Institute of Technology\\
  Hoboken, NJ 07030 \\
  \texttt{nwang7@stevens.edu} \\
  \And
 Mingxuan Chen \\
  Department of Electrical and Computer Engineering\\
  Stevens Institute of Technology\\
  Hoboken, NJ 07030 \\
  \texttt{mchen20@stevens.edu} \\
  \And
  K.P. Subbalakshmi \\
  Department of Electrical and Computer Engineering\\
  Stevens Institute of Technology\\
  Hoboken, NJ 07030 \\
  \texttt{ksubbala@stevens.edu} \\
}

\begin{document}
\maketitle
\begin{abstract}
\label{sec:abs}
In this work we propose three explainable deep learning architectures to automatically detect patients with Alzheimer’s disease based on their language abilities.
The architectures use: (1) only the part-of-speech features; (2) only language embedding features and (3) both of these feature classes via a unified architecture. 
We use self-attention mechanisms and interpretable 1-dimensional Convolutional Neural Network (CNN) to generate two types of explanations of the model's action: intra-class explanation and inter-class explanation. The inter-class explanation captures the relative importance of each of the different features in that class, while the inter-class explanation captures the relative importance between the classes.
\textit{Note that although we have considered two classes of features in this paper, the architecture is easily expandable to more classes because of its modularity}. Extensive experimentations and comparison with several recent models show that our method outperforms these methods with an accuracy of $92.2$\% and F1 score of $0.952$ on the DementiaBank dataset while being able to generate explanations. We show by examples, how to generate these explanations using attention values.
\end{abstract}

\section{Introduction}
In 2019, Americans spent \$244B in caring for patients with Alzheimer's Disease and Related Dementia (ADRD). The National Academy of Sciences, the National Plan to Address Alzheimer’s Disease, and the Affordable Care Act through the Medicare Annual Wellness, all identify earlier detection of ADRD as a core aim for improving the brain health for millions of Americans. The success of disease modification and preventive therapeutics for ADRD requires the identification of the disease in very early stages, at least a decade before onset. Approaches to early identification have included the use of brief cognitive screening tests and biological markers (usually neuroimaging or cerebrospinal fluid examination \cite{PitEtal98}). Neuroimaging modalities often include magnetic resonance imaging(MRI) \cite{KilEtal00} or the evaluation of positron emission tomography (PET) \cite{FosEtal83} targeting amyloid, Tau or both. The traditional biological marker methods tend to be invasive, expensive and create patient compliance problems. Hence, there is a strong motivation to consider early detection schemes using non-invasive markers of the disease.

In practice, cognitive assessment tools like Practitioner assessment of Cognition (GPCOG) \cite{BroEtal02} Cambridge Cognitive Examination (CAMCOG) \cite{SchEtal00}, Mini-Cog \cite{BorEtal00}, Mini-Mental State Examination (MMSE) \cite{GalEtal05} etc. are used to  
classify dementia and mild cognitive impairment (MCI) 
Even in primary care settings, cognitive impairment is unrecognized in $27$\%–$81$\% of the affected patients \cite{CorEtal13}. 
 
As language functions play an important role in the detection of cognitive deficiency across different stages of ADRD, speech transcripts can assist in early detection of the disease.
Hence techniques at the nexus of natural language processing and deep learning offer an inexpensive solution to this early detection problem \cite{BucEtal00,CroEtal96}. 

A Convolutional Neural Network (CNN) and Long Short Term Memory Network (LSTM) model (CNN-LSTM) was proposed in \cite{karEtal18, PalEtal19}, by using part-of-speech (PoS) tags to get accuracy up to $89.7$\% on classification for AD disease on DementiaBank cookie sub-corpus. 
Deep neural networks and deep language models were combined \cite{OriEtal18} to classifying the disease. On a sparse clinical language dataset, the model could predict MCI and AD type dementia with accuracy of $83$\%. 

One of the key criticisms of deep learning (DL) methods is 
that the decision process of the DL model is
intractable thereby making it difficult to understand the 
reasoning behind its decisions. For applications such
as AD detection, it is imperative that some form of reasoning 
be provided, because of the human angle involved. 
Hence in this work, \emph{we develop explainable deep learning architectures 
using attention mechanism and 1-dimensional CNN (1-D CNN) to detect Alzheimer's disease (AD) from 
transcripts of an individual's speech}.

The main contributions of our work are:
\begin{itemize} \itemsep 2pt
    \item novel explainable deep learning architectures for early detection of Alzheimer's disease
    \item inter and intra-feature attentions that captures the relative importance between the different feature classes as well as the relative importance of features within a class. These can be used to provide explanation of the model's conclusions
    \item modular architecture that can be extended to as many feature classes as desired
    \item extensive testing on the Dementiabank corpus, a popular Alzheimer's disease dataset
    \item example explanations using the proposed models
\end{itemize}
The rest of the paper is organized as follows.
Section~\ref{sec:related work}, describes related work. Section~\ref{sec:model and method}, discusses the proposed explainable deep learning models. Section~\ref{sec:experiment}, details the datasets, the experimental set up, data bias compensation mechanism, example explanations and discussions of the results. Section~\ref{sec:conlusion} discuses the conclusions we draw from this work.

\section{Related Work}
\label{sec:related work}
Automatically detecting Alzheimer's disease
using transcripts of conversations is not new.
Some previous efforts \cite{BarnEtal17,FraEtal16} used linguistic features, 
such as PoS tags and syntactic complexity; 
psycholinguistic features (e.g, lexical and morphological complexity, word information, and readability etc.) to detect Alzheimer's disease. \cite{FraEtal16} used regression models to achieve an accuracy of 81\% on classifying between AD patients and healthy controls.
Some researchers have combined latent features obtained via language 
embeddings like word2vec \cite{MikoEtal13} 
‘GloVe’ \cite{PennEtal14}
sentence2vector \cite{QuocEtal14} 
along with hand-crafted features in a 
\cite{MirEtal18,karEtal18,PalEtal19} hybrid approach using a CNN and RNN architecture to achieve accuracy around 89\%.




While the above methods show varying degrees of accuracy, they do not provide adequate explainability in their models.
Some researches have introduced visualizations to help move the research in the direction of explainability. Activation clusters and first derivative saliency heatmap were used in \cite{karEtal18}. K-means clustering was used \cite{JabEtal19} to find the most frequently used topic in the text data.
Note that explainable AD models have been developed for MRI based AD detection methods, such as \cite{OhEtal19} which proposed a gradient-based visualization method to detect the most important biomarkers related to AD and progressive mild cognitive impairment (pMCI).
However, these are not directly applicable to the language based detection problem that we are considering here.


Meanwhile, work on explainable AI has started to emerge in importance specifically to address the problem of trustablity of AI systems.
Multiple surveys have analyzed those works \cite{GilEtal18,ZhaEtal18,ChaEtal17} which focus on different areas of explainable AI (XAI).  \cite{ZhaEtal18} revisited visualization of CNNs and discussed trends in explainable artificial intelligence. \cite{ChaEtal17} categorized prior work into model transparency and model functionality on multitude dimensions for model interpretability and analyzed the insufficiency of current work.

An interpretable convolutional neural network (ICNN) was developed in \cite{ZhaEtal18} for use on images and tested on three benchmark image datasets. 
The structure of LSTM was explored and contribution of variables to the prediction in multi-variable time series was captured in \cite{GuoEtal19}. 
\cite{RibEtal16} developed local interpretable 
model-agnostic explanations (LIME) to explain both the 
predictions and models. 
\cite{ShrEtal17} presented a mechanism they call ``deep learning important features(DeepLIFT)" to specify the 
contributions of all neurons in the network to every feature of the input. 
Meijer G-functions were used to disclose the functional forms learned by a model without much apriori assumptions \cite{AlaEtal19}. 
However, none of these approaches provided precise explainable results and/or worked specifically in NLP domain. 

\textit{In this work, we adapt the multi-head self-attention (MHA) proposed in \cite{vasEtal17} and use 1-D CNN \cite{GolEtal16} to define two types of explanations: intra and inter-feature explanations. 
These 
capture the relative importance between features within a set
as well as between different class of features respectively. }

\section{Model and Method}
\label{sec:model and method}
%
We propose explainable AI models for detecting AD from speech transcripts of patients using attention mechanisms \cite{vasEtal17} and the 1-D CNN described in\cite{GolEtal16} to interpreting the decisions that the AI models make. 
We use two types of features in the model, the self-attention mechanism and a 1-D CNN to understand the relative importance of the features in the final outcome. 
There is a debate on whether attention mechanisms are good for interpretation \cite{JaiEtal19, WieEtal19}. 
However, this debate was settled in favor of using self-attention mechanisms as a viable method for interpretation for classification tasks \cite{VasEtal19}. 
Moreover, 1-D CNN's can be used to interpret an AI's decision as demonstrated in \cite{GolEtal16,JacEtal18} for NLP tasks.

\subsection{Proposed Architectures: C-Attention Networks}
\label{Sec:all-arc}
We propose three architectures: one that uses only PoS 
features, one that uses only the latent features (language embeddings) and a unified architecture, which uses both features.

\subsubsection{C-Attention-FT Network}
\label{sec:FT-net}
\begin{figure}[hbt!] 
\centering
\includegraphics[width=0.48\textwidth]{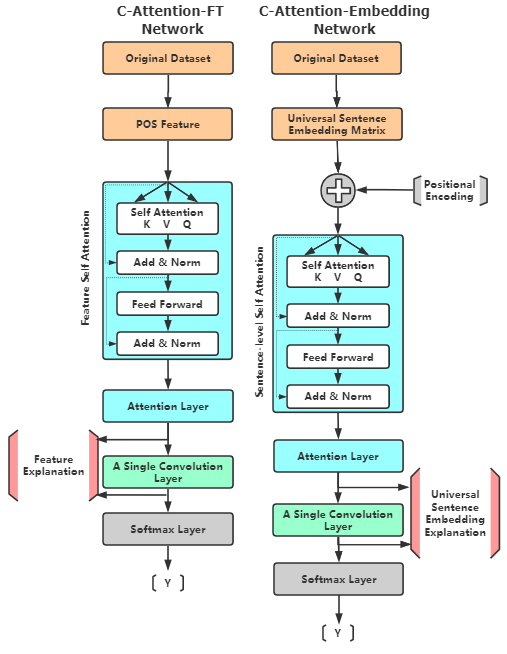}
\caption{The proposed architecture of C-Attention-FT Network and C-Attention-Embedding Network. The C-Attention-FT network uses the PoS features and the C-Attention-Embedding network uses the sentence embeddings of the patient/control's description.}
\label{fig:C_Attention_l}
\end{figure}

The architecture of the model proposed for exploiting PoS features (C-Attention-FT Network) is depicted on the left hand side of Figure~\ref{fig:C_Attention_l}.
This architecture comprises of a self-attention module that captures 
the intra-feature relationships; an attention layer together with a following 1-D CNN layer
that can be used 
to generate feature level explanations followed by 
a softmax layer. The MHA module is the same as that proposed in 
\cite{vasEtal17} for the popular transformer architecture and is 
presented in Sec~\ref{sec:attn}.
Let $R=\{r_1,r_2,...,r_n\}$ be the set of records, then $r_i$ is the $i^{\rm {th}}$ record in the dataset.  
We compute PoS tags for each record using NLTK \cite{FraEtal16}. Let $P=\{p_1,p_2,...,p_n\}$ be the set of
PoS feature vectors and $p_i$ be the $i^{\rm {th}}$ vector in the PoS matrix. 
We use $h$ Multi-Head-Attention (MHA)  layers on $P=\{p_1,p_2,...,p_n\}$ to capture the relationship between the PoS features. 
The MHA transforms $P=\{p_1,p_2,...,p_n\}$ to another matrix of  $n$-dimensional vectors $A=\{a_1,a_2,...,a_n\}$. 
The MHA module is followed by a 1-layer CNN and a softmax layer to get the final classification.

\subsubsection{C-Attention-Embedding Network}
\label{sec:emb-net}
The architecture of the proposed C-Attention-Embedding Network is shown on the right hand side of 
Figure~\ref{fig:C_Attention_l}. 
We propose this architecture as a means of 
capturing latent feature information implicit in language embeddings. 
Specifically we use the universal sentence embedding (USE) \cite{CerEtal18} to represent each sentence in a record.
This architecture is similar to the proposed
C-Attention (Sec~\ref{sec:FT-net} except 
for the addition of a positional encoding 
module. The positional encoding module is 
used to maintain the relative positions of
the sentences and is the same as that used in the transformer \cite{vasEtal17} architecture.

Let $u_i$ be the USE vector corresponding to the $i^{\rm {th}}$ 
sentence in the record. The positional encoding is applied to each 
vector and the resulting vectors,  $u'_i$, are used to construct the matrix
$U=\{u'_1,u'_2,...,u'_n\}$. An $h$-layer MHA module is used to
extract the intra-feature relationships in this architecture. 
This is followed by an attention layer that captures
interpretation at the embedding feature level. The output of the attention layer is fed to a $1$-layer CNN and a softmax to get the final prediction.

\subsubsection{Unified C-Attention Network}
\label{sec:unified-arch}
\begin{figure}[hbt!] 
\centering
\includegraphics[width=0.48\textwidth]{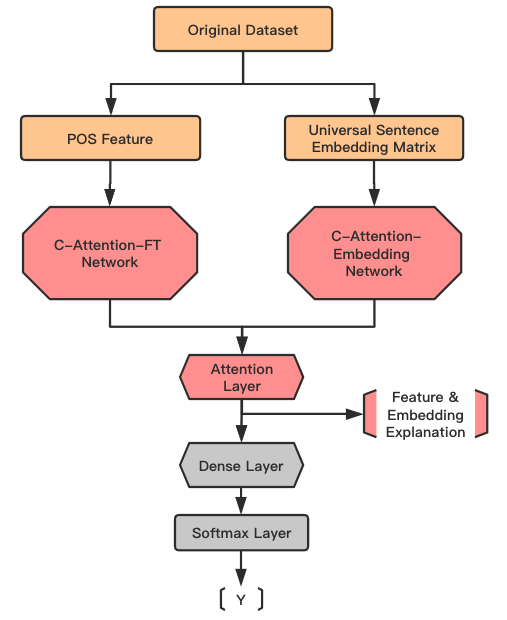}
\caption{The Architecture of Unified C-Attention Network for Feature and Embedding }
\label{fig:C_Attention}
\end{figure}

The third architecture we propose uses both the PoS and latent features of the sentences and is depicted in Figure~\ref{fig:C_Attention}. 
This architecture uses the proposed C-Attention-FT network and the C-Attention-Embedding network as two legs and combines them with another attention layer followed by a dense layer and the softmax layer. The dense linear layer is the same as that proposed in the transformer \cite{vasEtal17}. The attention layer captures the relative importance between the PoS and the USE features and helps in providing an interpretation at the feature class level.

\subsubsection{Attention Mechanisms}
\label{sec:attn}
%
Attention mechanisms have proved to be efficient in 
capturing global dependencies without any 
restrictions on the distance between the input and output sequences \cite{BahEtal14,YanEtal16}. 
Vaswani et al. \cite{vasEtal17} use self-attention 
\cite{ParEtal16} mechanisms along with positional coding in the design of the transformer which has become very popular in language models like BERT \cite{DevEtal18}. 
In this paper, we use the attention mechanisms and MHA mechanism proposed in \cite{vasEtal17}. 
These use a scaled dot product attention, which is given by 
\begin{equation}
\resizebox{.6\hsize}{!}{
 ${\rm Attention}(Q,K,V)= {\rm softmax} \left (\frac{QK^T}{\sqrt{{d}_k}} \right)V $}
 \end{equation}
 where $Q,K$ and $V$ are the query, key and value matrices and $d_k$ is dimension of the query and key vectors.
\subsubsection{1-D Convolution Layer}
\label{sec:cnn} 
We use a single layer 1-D CNN \cite{KorEtal18} as the penultimate layer, followed by a maxpooling layer which empowers the CNN to compress the information and extract global features. 
It was shown in \cite{GolEtal16} that 1-D CNNs filters essentially are
n-gram detectors with each filter specializing in a closely-related family of n-grams. This feature can be used to interpret the action
of the CNN layer in the architecture. We show how to trace back through these filters and the MHA layer to derive explanations for the classifiers outcomes.



\section{Experiment and Result}
\label{sec:experiment}

\begin{table*}[t]
\centering
\resizebox{\textwidth}{35mm}{
\begin{tabular}{cccccccccc}
\hline
 \textbf{Approach }& \textbf{Accuracy}  & \textbf{Precision } &\textbf{Recall }& \textbf{F1}  & \textbf{AUC } &\textbf{TN }& \textbf{FP}  & \textbf{FN } &\textbf{TP }   \\  
 \hline
  C-LSTM & 0.8384&0.8683&0.9497&0.9058&0.9057&6.3&15.6&5.3&102.6  \\
  C-LSTM-ATT&0.8333&0.8446&0.9778&0.9061&0.9126&2.6&19.3&2.3&105.6 \\
  C-LSTM-ATT-W&0.8512&0.9232&0.8949&0.9084&0.9139&14&8&11.3&96.6  \\
  C-BILSTM&0.8495&0.8508&0.9965&0.9178&0.9207&1&16.6&0.3&95  \\
  C-BILSTM-ATT&0.8466&0.8525&0.9895&0.9158&0.9503&1.3&16.3&1&94.3  \\
  C-BILSTM-ATT-W&0.882&0.9312&0.9298&0.9305&0.9498&11&6.6&6.6&88.6 \\
  C-BILSTM-ATT-W NO PSYCH.&0.879&0.887&0.9825&0.9319&0.9499&12&5.6&1.6&93.6 \\
  C-BILSTM-ATT-W NO SENT.&0.897&0.9239&0.9615&0.9321&0.9501&7.6&10&3.6&91.6 \\
  C-BILSTM-ATT-W NO DEMO.&0.8908&0.9005&0.9789&0.9308&0.9473&10.33&7.33&2&93.3\\

 \hline
 \textbf{Attention-FT}&0.868&0.895&0.94&0.917&0.924&18&11&6&94 \\
 \textbf{Attention-Embedding}&0.822&0.881&0.89&0.886&0.824&17&12&11&89 \\
 \textbf{Attention-FT+Embedding}&0.829&0.882&0.90&0.891&0.828&17&21&10&90 \\
 \textbf{C-Attention-FT}&\textbf{0.922}&0.935&0.971&\textbf{0.952}&0.971&19&7&3&100 \\
 \textbf{C-Attention-Embedding}&0.845&0.885&0.92&0.902&0.837&17&12&8&92 \\
 \textbf{C-Attention-FT+Embedding}&0.915&\textbf{0.969}&0.922&0.945&\textbf{0.977}&23&3&8&95 \\
  \hline  
\end{tabular}}
\caption{The comparison of performance metric between our models and others\' : C-LSTM, C-LSTM-ATT and C-LSTM-ATT-W were referenced from \cite{karEtal18}, their standard model kernel is Convolutional-LSTM; 
C-BILSTM, C-BILSTM-ATT, C-BILSTM-ATT-W, C-BILSTM-ATT-NO PSYCH., C-BILSTM-ATT-W NO SENT. and C-BILSTM-ATT-W NO DEMO. were referenced from \cite{PalEtal19}, the main kernel of these architectures is Convolutional-BILSTM. The rest six models are our models: Attention-FT is attention model with only PoS features; Attention-Embedding is the one with only universal sentence embedding; Attention-FT+Embedding is the combination of those two parts. C-Attention-FT, C-Attention-Embedding and C-Attention-FT+Embedding are the similar architectures but replaced the dense layer by a convolutional layer. All these six models are based on Attention Mechanism.
  }
\label{tab:results}
\end{table*}

We evaluate the proposed C-Attention Network 
architectures on the DementiaBank dataset and compare the performances
of these architectures with each other as well as some recently published results \cite{karEtal18, PalEtal19}.

\subsection{Data and Pre-processing}
\label{sec:data}
DementiaBank \cite{JamEtal94} is a database of multimedia interactions for the study of 
communications between people with dementia and a control group. The  patients group consists of individuals in the approximate age range of 49 to 90 years old while the control group has an approximate age range of 46 to 81 years. The dataset contains 
transcripts from four sub-tasks: (1) Cookie theft description: participants in both the control group and dementia group were given a picture of a 
child attempting to steal a cookie and asked to
describe what they saw; 
(2) Word fluency: which measures their language fluency (dementia group only); (3) Recall: measures the memory recall of the participants tested (dementia group only) and (4) Sentence construction: where they were tested on sentence construction (dementia group).
In total the corpus contains 1049 transcripts from 208 AD patients and 243 transcripts from 104 
elderly control individuals for a total of 1229 transcripts. We use this corpus for our experimental validation.


\subsection{Experiment Setup}
\label{sec:setup}
We implemented our proposed model by using Pytorch. The model is trained to minimize the cross-entropy 
loss function of predicting the class label of participants' speech records in the training set. As mentioned 
earlier, two types of features were extracted: part of speech (PoS) \cite{FraEtal16} and sentence embedding. We used the USE proposed in \cite{CerEtal18} for the embedding feature. For all models in our 
experiments, we have $6$ layers for the multi-head attention (MHA) module. We used 
stochastic gradient descent + momentum (SGD + Momentum) \cite{Rud16} as the optimizer for training.
Since the cookie theft sub-dataset is unbalanced we 
added a class weight correction by increasing the penalty for misclassifying the less frequent class 
during
model training to reduce the affect of data bias, as in \cite{PalEtal19}.
The class weight correction ratio used in this paper is $7:3$. 
The average number of utterances in this dataset is $17$.
In order to have a fixed number of utterances in our model, we set the number of utterances as $17$. 
We truncated extra utterances for descriptions that had more than $17$ utterances and added padding for those with less than $17$ utterances. 
Note that changing this number to the median
number of utterances or the maximum number of utterances did not give us better results. 
We randomly split the original data into 81\% training, 9\% validation and 10\% testing.
\begin{table*}[t]
\centering
\begin{tabular}{p{1.3cm}p{9.5cm}p{3.3cm}p{0.1cm}}
\hline
\hline
\textbf{Label }& \textbf{Speech Record}& \textbf{Important Sentences} \\
\hline
\hline

\multirow{6}{1cm}{0}&\multirow{6}{9cm}{okay, well the mother is drying the dishes, the sink is overflowing, um the little girl’s reaching for a cookie, and her brother’s taking cookies out of the cookie jar, and the stool is going to f knock him on the floor laughs, he’s going to fall on the floor because the stool’s not uh what, with gravity, whatever, uh the uh curtains are blowing I think, that’s all I can see} & um the little girl’s reaching for a cookie  \\
\cline{3-4}

~&~ & with gravity \\
\cline{3-4}
~&~ & he’s going to fall on the floor because the stool’s not uh what \\
\cline{3-4}
\hline


\multirow{9}{1cm}{1}&\multirow{9}{9cm}{I would like to have a lead pencil, the tree is blossoming, I hope my child doesn't hafta go to the hospital , I hope my child doesn't hafta go to the hospital, I shouldn't say that because we have a daughter who's pregnant, and I do want her to go to the hospital, okay then, this winter has been a very cold one, the doctor said I, I sat in the chair by a the doctor, brief, I'm not, I forgot to try make them brief, the bureau drawer stands open, ,} & I would like to have a lead pencil \\
\cline{3-4}
~&~ & ~& \\
~&~ & I shouldn't say that because we have a daughter who's pregnant \\

\cline{3-4}

~&~ & , , \\
~&~ & ~& \\

\hline
\multirow{3}{1cm}{1}&\multirow{3}{9cm}{uh the pencil is on the desk, , leafing, meaning the leaves are opening, , cold q exc, and winter q exc, last year was a cold winter} & and winter q exc  \\
\cline{3-4}
~&~ & , ,\\
\cline{3-4}
~&~ & cold q exc\\
\hline
\hline  
\end{tabular}
\caption{Samples of explanation for correct predictions of three speech records. We sort embedding sentences based on the attention value and show the top three sentences. The Label is the ground truth: 0 represents the healthy control; 1 represents the patient. Attention values means how much importance of these sentences are for prediction. 
  }
\label{tab:expla}
\end{table*}

\subsection{The Benchmarks}
\label{sec:results}
We compare the performance of our architectures with recently published results in terms of accuracy, precision, recall, F1 score, area under the curve (AUC). These comparisons are shown in Table~\ref{tab:results}. We also include total number of true negatives (TN), false positives (FP), false negatives (FN) and true positives (TP) for completion. 
In addition to these baseline architecture, we also compare performances with 
our architectures with a slight modification. All of these architectures are described below.
\begin{itemize}
\itemsep -2pt
    \item \textbf{Attention-FT:} The Attention-FT architecture is a slightly modified version of our proposed C-Attention-FT architecture. In this version, we replace the CNN with the dense linear network used in \cite{vasEtal17}. 
    \item \textbf{Attention-Embedding:} The Attention-Embedding architecture is a modification of the C-Attention-Embedding architecture. Just as in Attention-FT, we replace the CNN with a dense linear network.
    \item \textbf{Attention-FT+Embedding:} This is a variant of the proposed Unified C-Attention-FT+Embedding architecture, with the two ``legs" of the original architecture replaced by the Attention-FT and Attention-Embedding architectures described above.
    \item \textbf{C-LSTM:} The C-LSTM architecture consists of a CNN followed by an LSTM layer. PoS and word embedding features are used~\cite{PalEtal19}.
    \item \textbf{C-LSTM-ATT:} This model is the same as the CNN-LSTM, but followed by an attention layer. The additional attention layer was used to detect specific linguistic patterns related to dementia detection not for explanations. This work did not correct for data bias ~\cite{PalEtal19}.
    \item \textbf{C-LSTM-ATT-W:} This model is the same as the C-LSTAM-ATT, except the bias in the data is corrected by assigning class weights~\cite{PalEtal19}.
    \item \textbf{C-BILSTM:} The C-BILSTM architecture is built upon the C-LSTM architecture by using bi-directional LSTM instead of LSTM. This work used PoS, word embedding and other handcrafted features includes psycho-linguistic, average sentiment and demographic features~\cite{PalEtal19}.
    \item \textbf{C-BILSTM-ATT:} This model is the same as the C-BILSTM but followed by an attention layer. This work did not correct for data bias~\cite{PalEtal19}.
    \item \textbf{C-BILSTM-ATT-W:} This model is the same as the C-BILSTM-ATT, except the bias in the data is corrected by assigning class weights~\cite{PalEtal19}.
    \item \textbf{C-BILSTM-ATT-W NO PSYCH:} This model is the same as the C-BILSTM-ATT-W but without the psycho-linguistic features~\cite{PalEtal19}.
    \item \textbf{C-BILSTM-ATT-W NO SENT:} This model is the same as the C-BILSTM-ATT-W but without the average sentiment feature~\cite{PalEtal19}.
    \item \textbf{C-BILSTM-ATT-W NO DEMO:} This model is the same as the C-BILSTM-ATT-W but without the demographic features~\cite{PalEtal19}.
    \end{itemize}

\subsection{Performance Analysis}
\label{sec:perf-anal}
From the Table~\ref{tab:results} we see that the best overall performance is achieved by both the proposed C-Attention-FT and unified C-attention-FT+Embedding architecture. The C-Attention-FT architecture performs best in terms of accuracy and F1 scores, and the C-attention-FT+Embedding performs best in terms of precision and AUC.
Attention-Embedding does the worst in terms of accuracy, although its performance is not too bad in terms of precision recall and F1 scores. 
We also note that using attention only without a convolutional layer does not seem to result in the best performance as evidenced by the fact that all three modifications: Attention-FT, Attention-Embedding and Attention-FT+Embedding are not near the top in terms of most scores. Comparing only the Attention-FT and Attention-Embedding results, we note that the Attention-FT model performs better on all metrics, which seems to suggest that the PoS features (used in Attention-FT) may have more value than the latent features with it comes to using only attentions and no convolutional layers.

\subsection{Explainability Analysis}
The advantage of our model
is that we can interpret the classification process of the model. Specifically, for each case, we can explain 
whether the model considers the PoS features or the latent feature as more important in its decision. 
Similarly, we can also use the self-attention 
weights of to determine the
relative importance of utterances within a single picture description and the relative importance of the different PoS features in arriving at the final decision.
\subsubsection{Explaining the Universal Sentence Embedding Features:}
Table~\ref{tab:expla} shows some sample sentences with their corresponding ground truth (labels). Label 0 indicates that the description was uttered by a healthy individual (control) and Label 1 indicates that the description was uttered by a patient. In each of these examples the unified C-Attention-FT+Embedding network classified the utterance correctly. The column ``Important Sentences" refers to the sentences that are captured by the attention and 1-D CNN layers, which indicates higher importance of those sentences. 

According to the analysis on entire testing dataset (129 speeches), we have following findings: 
\begin{itemize}
\item the sentences that are considered most important by the attention layer (have highest attention values) is almost always captured by the filters in the 1-D CNN layer. We notice that 121 speeches out of 129 speeches show this pattern. 
\item We also note that the intra-feature attention value for a patient's utterances seems to be more uniformly distributed compared to that of a healthy control which shows a definite higher value for some sentences compared to others. This might indicate that the AI is picking up on the ``randomness" of the utterances of the patients compared to that of a health control.
\end{itemize}

\subsubsection{Explaining the role of Part of Speech Features:}
\begin{figure}[hbt!] 
\centering
\includegraphics[width=0.48\textwidth]{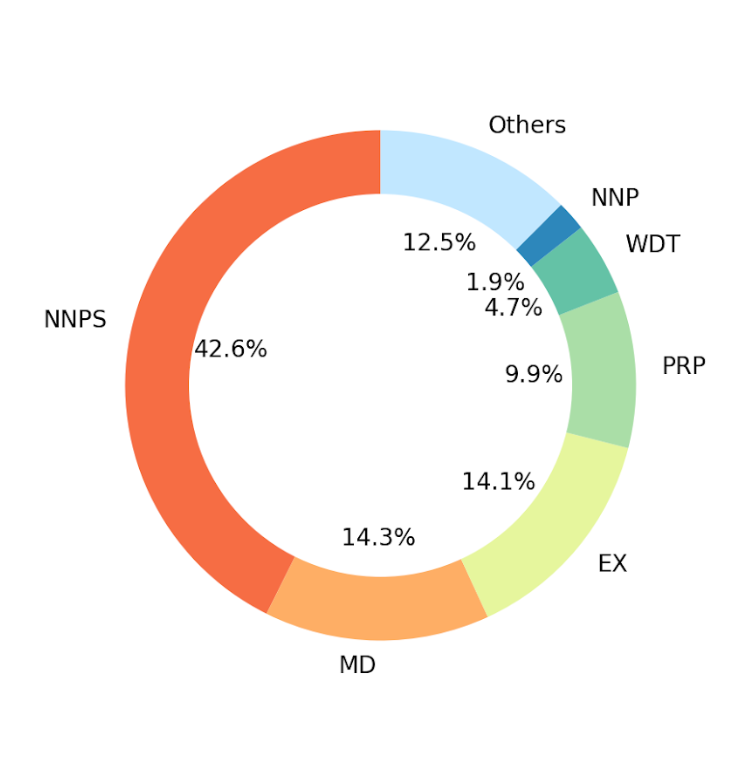}
\caption{The Top PoS Features Captured by the C-Attention-FT+Embedding to make the final decision. The result is obtained by analysing on the entire testing dataset, including 129 speeches.}
\label{fig:POS}
\end{figure}

PoS features are extracted at speech level. A total of $36$ PoS tags are defined. Like the
latent features we notice that the PoS features
with the highest attention weight are 100\% 
captured by the the filters in the following 
1-D CNN layer.
The top PoS features captured by attention layer and 1-D convolutional layers are shown in Figure~\ref{fig:POS}. 
This result is obtained by analysing the entire testing dataset. 
It would indicate that the PoS features NNPS (proper noun, plural), MD (modal), EX (existential) and PRP (personal pronoun) are 
the most important accounting for $80.9\%$ of times for the PoS Features. Our findings are consistent with previous works \cite{JaEtal14} that compared showed that AD patients tend to use more pronouns instead of nouns compared with healthy controls. 


\subsubsection{Explaining the relative importance of feature classes (PoS vs latent features):}
\begin{figure}[hbt!] 
\centering
\includegraphics[width=0.48\textwidth]{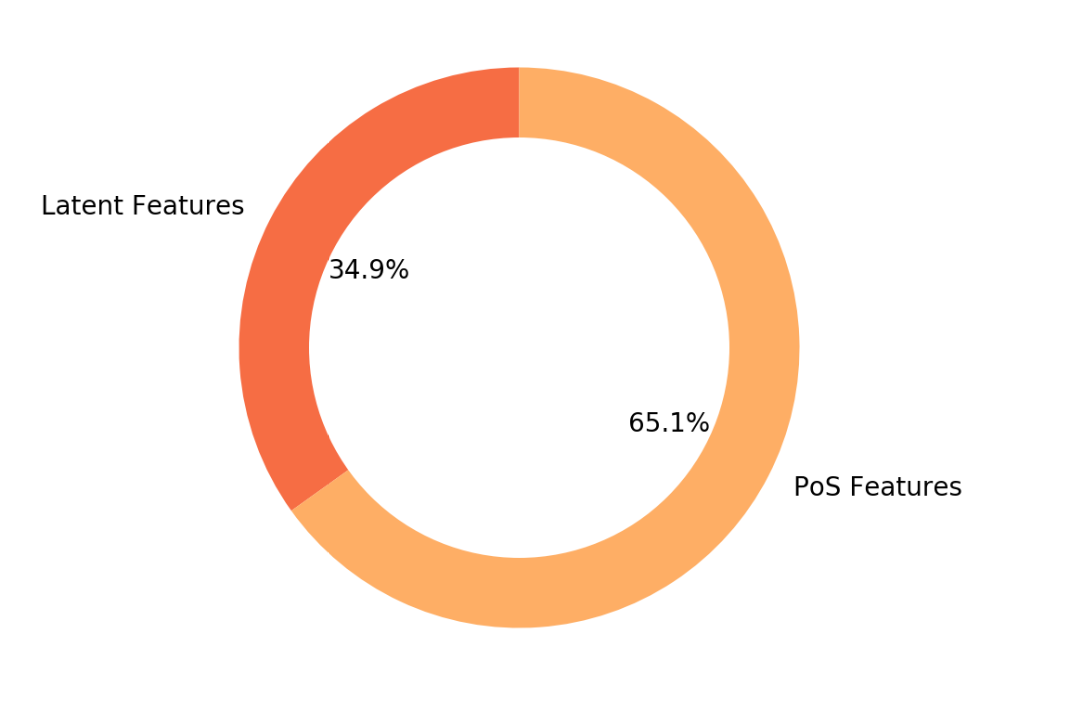}
\caption{Attention value for Part of Speech leg  is higher than Universal Sentence Embedding in 65.1\% of cases over the entire dataset.}
\label{fig:attAll}
\end{figure}

The final attention layer of the proposed unified C-Attention-FT+Embedding architecture (Figure~\ref{fig:C_Attention}), captures the relative importance between the PoS features and the latent features in a decision. 
Table~\ref{tab:dense} shows these attention values for three patients (with their record numbers indicated). Record number 1188 and 982 correspond to AD patients that were correctly identified as patients by the C-Attention-FT+Embedding model. Record 182 corresponds to a healthy control (indicated by a label value of 0) and also correctly identified by the model.
For example, for speech record number 1188, 
the latent features played a bigger role in determining the overall decision with a weight of $0.703$.  By contrast, for record number 182, the PoS feature seems to have weighed slightly more ($0.572$) than the latent features ($0.428$).
Figure~\ref{fig:attAll} shows that the "Part of Speech " leg is assigned a higher attention value in 65.1\% of the cases, while Universal Sentence Embedding is assigned a higher attention value in 34.9\% of the cases, indicating that the PoS features seem to play a higher role in detecting AD. This fact is also demonstrated in Table~\ref{tab:results}.

\begin{table}[ht!]
\begin{center}
\begin{tabular}{ |c|c|c|c| } 
\hline
\textbf{Tag} & \textbf{Meaning} & \textbf{Example} \\ 
\hline
NNPS & proper noun, plural  & Americans \\
\hline
MD & modal could & will \\
\hline
EX\$ & existential there & there is \\
\hline
PRP & personal pronoun & I \\
\hline
WDT & wh-determiner & which \\
\hline
NNP\$ & proper noun, singular & Harrisons \\
\hline
\end{tabular}
\caption{Example of PoS tags}
\label{tab:tag}
\end{center}
\end{table}
\begin{table}[hbt!]
\centering
\setlength{\abovecaptionskip}{0pt}%
\setlength{\belowcaptionskip}{10pt}%

\label{tab:baselines}
\begin{tabular}{p{1cm}p{1.5cm}p{1.8cm}p{1.8cm}} 
\hline
  \textbf{Label} & \textbf{Record Number} &\textbf{PoS Weight } & \textbf{Embedding Weight}   \\  
 \hline
1&  1188 & 0.297  & 0.703  \\

  0&182  & 0.572  & 0.428  \\

  1&982 & 0.157  & 0.843  \\

  \hline
\end{tabular}
\caption{Attention Values for Part of Speech and Universal Sentence Embedding for three speech records numbered 1188, 182 and 982. Label 0 corresponds to a healthy control and Label 1 corresponds to an AD patient. In all cases, the C-Attention-FT+Embedding model identified the status of the individual correctly.}
\label{tab:dense}
\end{table}

\section{Conclusion}
\label{sec:conlusion}
We proposed three explainable architectures using CNN 
and attention to detect Alzheimer's disease using two kinds of features: part-of-speech and language embeddings. One architecture uses only the PoS feature, one uses only the universal sentence embedding and the third is a unified architecture that uses both of these features. We propose the use of attention layers and 1-D CNN layer to capture explanations at 2 levels: one each at the intra-feature level 
and inter-feature-class level. The 
intra-feature level attention weights and 1-D CNN filters capture the relative importance the 
model places on the individual features in the category, whereas the 
inter-feature level attention weights gives us an idea of the relative 
importance that the model placed between the two classes of features.

Extensive testing on the popular DementiaBank datasets and comparisons with
several recently published models as well as minor modifications of our own
models show that \textbf{the C-Attention-FT architecture performs best in terms of accuracy and F1 scores, and the C-attention-FT+Embedding performs best in terms of precision and AUC while at the same time being able to generate explanations of the action of the AI}. 
We also show by examples how to generate explanations for the actions of the models. 
Our results agree with some of the previous work that shows that AD patient's tend to use more 
pronouns instead of nouns. 
Our work thus is an inexpensive, non-invasive, explainable
AI model that can detect AD at good performance metric. Since it is based on only the spoken language, it can be potentially easily implemented in an app setting there by giving the option of taking it at home. This in turn
can have a positive impact on patient compliance and 
therefore early detection of AD. 







\bibliography{kdd_ws_2020}
\bibliographystyle{unsrt}
\end{document}